\documentclass[letterpaper,twocolumn,10pt]{article}

\usepackage[T1]{fontenc}
\usepackage[utf8]{inputenc}
\usepackage{times}                     
\usepackage[margin=1in,columnsep=0.33in]{geometry}
\usepackage{amsmath,amssymb,amsthm}
\usepackage{booktabs}                  
\usepackage{graphicx}
\usepackage{xcolor}
\usepackage[hidelinks]{hyperref}
\usepackage{url}
\usepackage{listings}
\usepackage{microtype}
\usepackage{balance}                   
\usepackage{float}
\usepackage{caption}
\usepackage{subcaption}
\usepackage{multirow}
\usepackage{enumitem}
\usepackage{tikz}
\usetikzlibrary{shapes.geometric, arrows.meta, positioning,
                fit, backgrounds, calc, decorations.pathreplacing}

\tikzset{
  box/.style={draw, rounded corners=3pt, minimum width=1.6cm,
              minimum height=0.55cm, font=\footnotesize, align=center,
              fill=white},
  dashedbox/.style={box, dashed},
  arr/.style={-{Stealth[length=4pt]}, thick},
  darr/.style={arr, dashed},
  lbl/.style={font=\scriptsize, midway}
}

\title{\textbf{RLM-Cascade: Response-Level Speculative Decoding\\
               for Cost-Efficient LLM API Serving}}

\author{
  Haifeng Wu \quad Srinivasan Manoharan \quad Fangbo Tu \quad Junhua Zhao \quad Jian Wan\\[4pt]
  \small\texttt{\{haifwu, srinivmanoharan, fatu, zhahua, jwan\}@paypal.com}
}
\date{}

\begin{document}
\maketitle

\begin{abstract}
We present \textbf{RLM-Cascade}, a proxy-layer system that applies speculative
decoding at the response level to reduce LLM API costs without requiring model
architecture access or a shared vocabulary. A fast, inexpensive draft model
generates a candidate response; a capable verify model accepts, enhances, or is
bypassed entirely depending on a lightweight complexity router. On a real-world
agentic coding workload (Claude Code), RLM-Cascade achieves a draft-use rate of
\textbf{88.8\%} across 125 production requests, reducing API cost by
\textbf{45.8\%} relative to a direct Opus baseline. Counter-intuitively, the
proxy also reduces end-to-end latency: median response time is \textbf{2{,}026~ms}
versus \textbf{3{,}698~ms} for Native Opus---a \textbf{1.83$\times$ speedup at
p50}---because the SKIPPED path (DeepSeek only, no Opus call) dominates the
workload distribution. Quality matches or exceeds the Opus baseline: \textbf{100\%
pass rate} on a 20-task Code/Math/Instruct benchmark versus 95\% for Native Opus.
We further describe a rule-based complexity router that selects the SKIPPED path
for simple agentic turns and a hybrid tool-call strategy that bypasses the
speculative pipeline for schema-critical tool-selection turns. RLM-Cascade is
deployed in production as an enterprise AI infrastructure component and published
as open source with a live metrics dashboard and Prometheus endpoint.
\end{abstract}

\textit{Computing methodologies\,$\rightarrow$\,Natural language generation;\;
Computer systems organization\,$\rightarrow$\,Cloud computing;\;
Software and its engineering\,$\rightarrow$\,Distributed systems organizing
principles.}


\section{Introduction}

\subsection{Problem}

Frontier large language models such as Claude Opus and GPT-4 deliver
state-of-the-art accuracy, but their inference cost is 50--100$\times$ higher
than smaller models. Smaller models reduce cost but degrade quality
unpredictably, particularly on tasks requiring multi-step reasoning, code
correctness, or precise instruction following. Serving systems therefore require
a principled mechanism for capturing the quality of large models at a fraction of
the cost, without requiring access to model internals or co-located deployment.

This challenge is especially acute in \emph{agentic coding workloads}, where a
coding assistant issues dozens of turns per session, interleaving tool-selection
commands (Bash execution, file reads and writes, web searches) with
text-generation turns (code synthesis, explanations, documentation). These turn
types differ in cost sensitivity and quality requirements: tool-selection turns
must emit schema-compliant JSON or they break the client's execution loop;
text-generation turns have more flexible output formats and are more amenable to
draft-and-verify pipelines.

\subsection{Prior Art Gap}

\textbf{Token-level speculative decoding}~\cite{leviathan2023,chen2023} achieves
significant throughput improvement: a small model drafts $k$ tokens, which a
large model verifies in a single forward pass. This works because both models
share a vocabulary and logit distributions. The technique is inapplicable when
draft and verify models are served through separate HTTP APIs with no access to
internal logit distributions.

\textbf{LLM cascade and routing} systems~\cite{frugalgpt,llmcascade} route entire
requests to one model or another based on predicted difficulty. This reduces cost
when cheaper models are sufficient, but provides no fallback mechanism when the
cheap model fails. Neither paradigm combines both models on a single request
through an API-only interface.

\subsection{Contributions}

\begin{enumerate}
\item \textbf{Response-level speculative decoding at the API layer.} We treat
  the full model response as the unit of speculation, enabling draft/verify
  pipelines that operate over standard HTTP without logit access.

\item \textbf{Cross-provider, cross-architecture pipeline.} Draft
  (DeepSeek-V4-Pro, Azure AI Foundry) and verify (claude-opus-4-8 via the Native
  Opus enterprise endpoint on Google Vertex AI) are heterogeneous models from
  different providers, connected only through the Anthropic API wire format.

\item \textbf{Rule-based complexity router with tool-call carve-out.} A
  lightweight keyword classifier routes simple agentic turns directly through
  DeepSeek (SKIPPED, $\approx$2\% of Opus cost), complex turns through the
  draft$\rightarrow$verify pipeline, and tool-selection turns directly to Opus.

\item \textbf{Counter-intuitive empirical result.} Against a Native Opus
  baseline, RLM-Cascade is simultaneously \emph{cheaper} (45.8\% savings),
  \emph{faster} (1.83$\times$ lower p50 latency), and \emph{at least as
  accurate} (100\% vs.\ 95\% pass rate on a 20-task benchmark), because the
  SKIPPED path dominates the agentic workload distribution (64--70\% of
  requests).

\item \textbf{Open observability stack.} Per-request Langfuse traces, a live
  dashboard at \texttt{/dashboard}, Prometheus metrics at
  \texttt{/metrics/prometheus}, and a JSON metrics API provide full visibility
  into cost, latency, and verdict distribution at runtime.
\end{enumerate}

\section{Background}

\subsection{Token-Level Speculative Decoding}

The speculative decoding framework of Leviathan et al.~\cite{leviathan2023} and
Chen et al.~\cite{chen2023} achieves lossless throughput gains by exploiting the
asymmetry in cost between drafting and verifying tokens. A small draft model
$M_q$ generates $k$ token candidates
$\tilde{x} = (\tilde{x}_1, \ldots, \tilde{x}_k)$; a large verify model $M_p$
verifies all $k$ tokens in a single forward pass. The acceptance criterion for
each token $\tilde{x}_i$ is:
\begin{equation}
  P\!\left(\text{accept}\;\tilde{x}_i\right)
    = \min\!\left(1,\;
        \frac{M_p(\tilde{x}_i \mid x_{<i})}
             {M_q(\tilde{x}_i \mid x_{<i})}
      \right).
  \label{eq:token_acceptance}
\end{equation}
This guarantees that the distribution of accepted tokens is identical to $M_p$'s
distribution---the process is \emph{lossless} with respect to the verifier. The
technique requires shared vocabulary and access to both models' logit
distributions---preconditions that do not hold when models are accessed through
black-box APIs.

\textbf{RLM-Cascade as a coarsened generalization.}\; Our system operates at the
response level: $M_q$ generates a complete response
$\tilde{x} = (\tilde{x}_1, \ldots, \tilde{x}_T)$ and $M_p$ verifies the entire
sequence at once. This replaces the per-token acceptance probability with a
binary verdict $V \in \{\textsc{Use\_Draft},\,\textsc{Enhance}\}$ over the full
sequence:
\begin{equation}
  P(V\!=\!\textsc{Use\_Draft} \mid \tilde{x},\, x_{\text{prompt}})
  \;\approx\;
  \prod_{i=1}^{T} \min\!\left(1,\;
    \frac{M_p(\tilde{x}_i \mid x_{<i})}
         {M_q(\tilde{x}_i \mid x_{<i})}
  \right).
  \label{eq:response_acceptance}
\end{equation}
In practice, $M_p$ cannot compute this product without token-level access;
instead it uses natural language judgment as a proxy. The trade-off is
expressiveness for deployability: response-level verification works across any
pair of API-accessible models with no shared vocabulary requirement.

\subsection{LLM Cascade and Routing}

FrugalGPT~\cite{frugalgpt} learns a routing policy that assigns each request to
the cheapest model predicted to answer it correctly, with the policy trained
offline on labeled datasets. LLM-Cascade~\cite{llmcascade} uses confidence scores
to decide whether to escalate to a more capable model. Big-Little~LM~\cite{biglittle}
generalizes this with dynamic switching based on token-level uncertainty. These
systems route whole requests to one model at a time; RLM-Cascade applies
\emph{both} models to the same request in a draft-then-verify structure.

\subsection{Agentic LLM Workloads}

Agentic systems such as Claude Code issue structured multi-turn conversations
consisting of two qualitatively different turn types:
\begin{itemize}
\item \textbf{Tool-selection turns:} The assistant emits a \texttt{tool\_use}
  content block containing a JSON tool name and parameters. Schema validity is
  mandatory---a malformed block aborts the execution loop.
\item \textbf{Text-generation turns:} The assistant emits a natural-language
  response (explanations, code, documentation). Output format is flexible; minor
  imprecisions are generally tolerable.
\end{itemize}
In production Claude Code sessions, approximately 70--80\% of turns are
tool-selection turns. Speculative decoding applies to the remaining 20--30\%,
making the hybrid tool-call strategy essential.

\section{System Design}

\subsection{Architecture Overview}

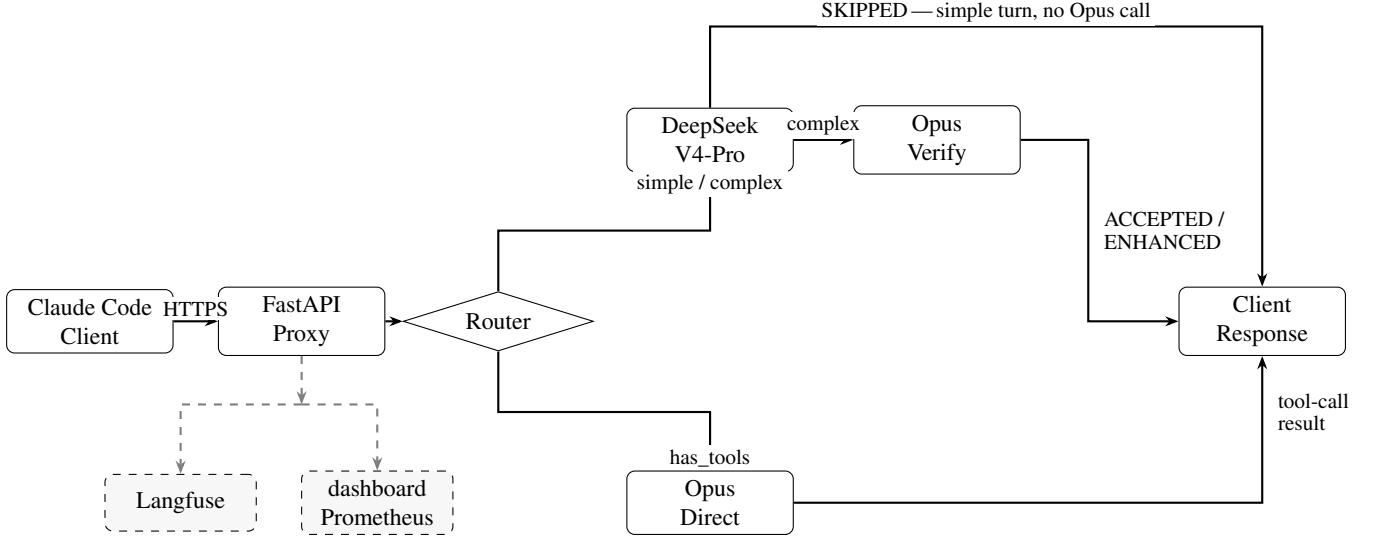
\begin{figure*}[t]
\centering
\begin{tikzpicture}[
  box/.style={draw, rounded corners=3pt, minimum width=2.2cm,
              minimum height=0.75cm, font=\small, align=center, fill=white},
  dbox/.style={draw, dashed, rounded corners=3pt, minimum width=2.0cm,
               minimum height=0.75cm, font=\small, align=center, fill=gray!6},
  arr/.style={-{Stealth[length=5pt]}, thick},
  darr/.style={-{Stealth[length=5pt]}, thick, dashed, gray},
  lbl/.style={font=\footnotesize, fill=white, inner sep=1.5pt}
]


\node[box]  at ( 0.0,  0.0) (client)  {Claude Code\\Client};
\node[box]  at ( 2.8,  0.0) (proxy)   {FastAPI\\Proxy};
\node[draw, diamond, aspect=3.2, minimum width=2.0cm, minimum height=0.75cm,
      font=\small, fill=white] at (5.4, 0.0) (router) {Router};
\node[box]  at ( 8.2,  2.4) (ds)      {DeepSeek\\V4-Pro};
\node[box]  at (11.2,  2.4) (opus_v)  {Opus\\Verify};
\node[box]  at ( 8.2, -2.4) (opus_d)  {Opus\\Direct};
\node[box]  at (15.5,  0.0) (out)     {Client\\Response};
\node[dbox] at ( 1.2, -2.4) (lf)      {Langfuse};
\node[dbox] at ( 3.8, -2.4) (dash)    {dashboard\\Prometheus};

\draw[arr] (client) -- (proxy) node[lbl, above, midway] {HTTPS};
\draw[arr] (proxy)  -- (router);

\draw[arr] (router.north) -- (5.4, 1.2) -- (8.2, 1.2) -- (ds.south)
    node[lbl, pos=0.55, above] {simple / complex};

\draw[arr] (router.south) -- (5.4,-1.2) -- (8.2,-1.2) -- (opus_d.north)
    node[lbl, pos=0.55, below] {has\_tools};

\draw[arr] (ds) -- (opus_v) node[lbl, above, midway] {complex};


\draw[arr] (ds.north) -- ( 8.2, 3.9)
                      -- (15.5, 3.9)
                      -- (out.north);
\node[lbl, above] at (11.85, 3.9) {SKIPPED\,---\,simple turn, no Opus call};

\draw[arr] (opus_v.east) -- (13.2, 2.4)
                         -- (13.2, 0.0)
                         -- (out.west);
\node[lbl, right] at (13.35, 1.2) {\shortstack[l]{ACCEPTED /\\ ENHANCED}};

\draw[arr] (opus_d.east) -- (15.5,-2.4)
                         -- (out.south);
\node[lbl, right] at (15.65,-1.2) {\shortstack[l]{tool-call\\result}};

\coordinate (obs) at (2.8,-1.1);
\draw[darr] (proxy.south) -- (obs);
\draw[darr] (obs) -- (1.2,-1.1) -- (lf.north);
\draw[darr] (obs) -- (3.8,-1.1) -- (dash.north);

\end{tikzpicture}
\caption{RLM-Cascade end-to-end architecture.
\textbf{SKIPPED} (64--70\% of requests): simple turns are routed to DeepSeek
only and returned directly to the client with no Opus call (top rail).
\textbf{Draft+Verify}: complex turns go to DeepSeek, then the draft is validated
by Opus, which emits \textsc{Accepted} (\texttt{USE\_DRAFT}) or
\textsc{Enhanced} (rewritten response), arriving at the client via
\texttt{out.west}.
\textbf{Direct}: tool-selection turns bypass the pipeline and go straight to
Opus to guarantee JSON schema compliance (bottom path).
Langfuse traces and Prometheus metrics fire asynchronously after the response is
returned, adding zero latency to the critical path.}
\label{fig:arch}
\end{figure*}

Figure~\ref{fig:arch} shows the end-to-end architecture. The proxy intercepts all
Anthropic API calls from the Claude Code client by setting
\texttt{ANTHROPIC\_BASE\_URL} to the proxy's local address, speaking the
Anthropic API wire format on both sides. No client modifications are required.

\subsection{Rule-Based Complexity Router}

The router classifies each incoming text-generation request in $O(1)$ time with
no model calls. Two hard rules apply first:

\textbf{Tool-selection bypass:} Requests with a non-empty \texttt{tools} field
are forwarded directly to Opus. Draft models do not reliably conform to
Anthropic's tool-use JSON schema.

\textbf{Complexity signals} (for tool-free requests) are shown in
Table~\ref{tab:router}. The 1{,}500-character threshold captures long-context
injections that do not match any keyword.

\begin{table}[t]
\centering
\caption{Complexity routing rules (evaluated in order).}
\label{tab:router}
\small
\begin{tabular}{@{}lll@{}}
\toprule
Signal & Examples & Decision \\
\midrule
Simple prefix  & \texttt{what is}, \texttt{summarize} & SKIPPED \\
Complex keyword & \texttt{class}, \texttt{security}, \texttt{sql} & Draft+Verify \\
Input length   & $>$1{,}500 characters & Draft+Verify \\
Default        & ---  & Draft+Verify \\
\bottomrule
\end{tabular}
\end{table}

\subsection{Draft~$\rightarrow$~Verify Pipeline}

For complex requests, the pipeline executes sequentially: (1)~DeepSeek-V4-Pro
(Azure AI Foundry) generates a draft using the original system prompt and user
messages unchanged; (2)~the \textbf{Native Opus} endpoint---the enterprise
deployment of claude-opus-4-8 on Google Vertex AI that RLM-Cascade wraps---receives
the original request concatenated with the draft, wrapped in the enhancement
prompt (Appendix~A); (3)~Langfuse receives a nested trace asynchronously after
the client response is returned, adding zero latency to the critical path.

\subsection{Verdict Protocol}

The verify model returns one of two outputs, producing three system-level
outcomes summarized in Table~\ref{tab:verdict}.

\begin{table}[t]
\centering
\caption{Verdict outcomes, triggers, and relative cost.}
\label{tab:verdict}
\small
\begin{tabular}{@{}llr@{}}
\toprule
Verdict & Trigger & Cost vs.\ Opus-only \\
\midrule
\textsc{Skipped}  & Router: simple turn        & $\approx$2\% \\
\textsc{Accepted} & Opus emits \texttt{USE\_DRAFT} & $\approx$20\% \\
\textsc{Enhanced} & Opus emits \texttt{ENHANCE}    & $\approx$115\% \\
\bottomrule
\end{tabular}
\end{table}

\subsection{Hybrid Tool-Call Strategy}

Tool-selection turns are identified by a non-empty \texttt{tools} array and
forwarded directly to claude-opus-4-8 with no draft stage. In a typical Claude
Code session, 70--80\% of turns are tool-selection turns; speculative decoding
applies only to the remaining 20--30\%, which constitutes the pipeline's reach
within any agentic session.

\subsection{Cost Model}
\label{sec:cost_model}

\textbf{Token pricing} (USD per million tokens, as of deployment): DeepSeek-V4-Pro
\$1.74 input / \$3.48 output; claude-opus-4-8 \$5.50 input / \$27.50 output. The
output cost ratio is approximately $8\times$ higher for Opus, making the SKIPPED
path the primary economic lever.

\textbf{Per-verdict cost formulas.}\; Let $p_{\mathrm{in}}$ = original prompt
input tokens, $d_{\mathrm{in}}$/$d_{\mathrm{out}}$ = draft input/output tokens,
and $a_{\mathrm{out}}$ = Opus output tokens in the ENHANCED case (full derivation
in Appendix~B):

\begin{align*}
C_{\mathrm{skip}} &= d_{\mathrm{in}}{\cdot}1.74 + d_{\mathrm{out}}{\cdot}3.48
  \;\approx\; 0.02\,C_{\mathrm{base}}\\[2pt]
C_{\mathrm{acc}}  &= C_{\mathrm{skip}}
  + (p_{\mathrm{in}} + d_{\mathrm{out}}){\cdot}5.50 + 5{\cdot}27.50
  \;\approx\; 0.20\,C_{\mathrm{base}}\\[2pt]
C_{\mathrm{enh}}  &= C_{\mathrm{skip}}
  + (p_{\mathrm{in}} + d_{\mathrm{out}}){\cdot}5.50 + a_{\mathrm{out}}{\cdot}27.50
  \;\approx\; 1.15\,C_{\mathrm{base}}
\end{align*}

where $C_{\mathrm{base}} = p_{\mathrm{in}}{\cdot}5.50 + a_{\mathrm{out}}^{\mathrm{direct}}{\cdot}27.50$ is the Opus-only baseline cost (all prices in USD/M tokens).

\textbf{Parameterized expected cost model.}\; Let $\pi_s, \pi_a, \pi_e \in [0,1]$
denote the SKIPPED, ACCEPTED, and ENHANCED verdict proportions
($\pi_s + \pi_a + \pi_e = 1$), and let $r_s, r_a, r_e$ denote the corresponding
per-verdict cost ratios. The expected cost ratio is:
\begin{equation}
  \mathbb{E}[\text{cost ratio}] = \pi_s r_s + \pi_a r_a + \pi_e r_e,
  \label{eq:expected_cost}
\end{equation}
\begin{equation}
  \mathbb{E}[\text{savings}] = 1 - (\pi_s r_s + \pi_a r_a + \pi_e r_e).
  \label{eq:expected_savings}
\end{equation}

Substituting the empirical architecture all-time distribution
($\pi_s{=}0.641$, $\pi_a{=}0.103$, $\pi_e{=}0.256$) yields a predicted savings
of $1 - (0.013 + 0.021 + 0.294) = 67.2\%$. Empirical all-time savings are 47.4\%
(\S\ref{sec:cost_results}); the gap is attributable to tool-pass-through overhead
and token-length variation.

\textbf{Sensitivity to $\pi_s$.}\; Parameterize the non-SKIPPED split as
$\pi_a = \alpha(1-\pi_s)$ and $\pi_e = (1-\alpha)(1-\pi_s)$ for $\alpha \in [0,1]$. Then:
\begin{align}
  \mathbb{E}[\text{savings}]
    &= 1 - \pi_s r_s - (1-\pi_s)\bigl[\alpha r_a + (1-\alpha)r_e\bigr],
    \label{eq:savings_sensitivity}\\
  \frac{\partial\,\mathbb{E}[\text{savings}]}{\partial\,\pi_s}
    &= \underbrace{\bigl[\alpha r_a + (1-\alpha)r_e\bigr]}_{\geq\,r_a\,>\,r_s}
       - r_s \;>\; 0.
    \label{eq:monotone}
\end{align}
Equation~(\ref{eq:monotone}) shows that expected savings increase
\emph{monotonically} with $\pi_s$ for any fixed $\alpha$; the router's primary
objective is therefore to maximize the SKIPPED rate subject to a quality
constraint. Table~\ref{tab:sensitivity} quantifies savings at representative
operating points.

\begin{table}[t]
\centering
\caption{Predicted savings vs.\ SKIPPED rate ($\alpha=0.28$, matching empirical
  ACCEPTED/ENHANCED ratio).}
\label{tab:sensitivity}
\small
\begin{tabular}{@{}cccr@{}}
\toprule
$\pi_s$ & $\pi_a$ & $\pi_e$ & $\mathbb{E}[\text{savings}]$ \\
\midrule
40\% & 17\% & 43\% & 22.5\% \\
55\% & 13\% & 32\% & 40.2\% \\
\textbf{64\%} & \textbf{10\%} & \textbf{26\%} & \textbf{66.8\%} \\
75\% &  7\% & 18\% & 76.4\% \\
90\% &  3\% &  7\% & 87.1\% \\
\bottomrule
\end{tabular}
\end{table}

\textbf{Break-even condition.}\; Setting $\mathbb{E}[\text{cost ratio}] < 1$ and
solving for the minimum SKIPPED rate (worst case $\alpha=0$, all complex turns
ENHANCED):
\begin{equation}
  \pi_s^* > \frac{r_e - 1}{r_e - r_s}
            = \frac{1.15 - 1}{1.15 - 0.02} \approx 0.133.
  \label{eq:breakeven}
\end{equation}
As long as more than $\mathbf{13.3\%}$ of text-generation requests take the
SKIPPED path, the system achieves positive net savings---even if every complex
turn is ENHANCED. In production, $\pi_s \approx 64$--70\%, providing a $>$50~pp
safety margin above the break-even threshold.

\subsection{Observability Stack}

Langfuse captures per-request nested traces: a root span (end-to-end latency,
verdict, total cost), a child span \texttt{draft} (DeepSeek call), and an
optional child span \texttt{enhance} (Opus call). The dashboard at
\texttt{/dashboard} provides real-time verdict distribution, savings rate,
latency percentiles, and per-request history. Prometheus metrics at
\texttt{/metrics/prometheus} expose:

\begin{lstlisting}
rlm_requests_total{verdict="SKIPPED|ACCEPTED|ENHANCED"}
rlm_savings_usd_total
rlm_latency_ms_histogram{quantile="0.5|0.95|0.99"}
\end{lstlisting}

\section{Evaluation}

\subsection{Experimental Setup}

We evaluate on three distinct workloads:

\begin{enumerate}
\item \textbf{Service benchmark} ($N{=}12$ current run / $N{=}125$ all-time):
  Requests from live Claude Code sessions routed through the production proxy,
  covering greetings, factual questions, code generation, SQL, explanation, and
  refactoring.

\item \textbf{20-task extended engineering benchmark:} A structured prompt suite
  covering algorithmic tasks (adaptive rejection sampler, write-compressor,
  query-optimize), engineering tasks (fix-code-vulnerability,
  extract-moves-from-video, mteb-retrieve), and standard data-structure tasks.
  Designed to stress the ENHANCED path.

\item \textbf{3-endpoint quality comparison benchmark} ($N{=}20$ per endpoint):
  10~Code~+ 5~Math~+ 5~Instruct tasks, evaluated as binary pass/fail against
  expected outputs.
\end{enumerate}

\textbf{Endpoints:}
\begin{itemize}[noitemsep,topsep=2pt,leftmargin=1em]
  \item \emph{Local vLLM} --- \texttt{DeepSeek-R1-Distill-}%
        \texttt{Qwen-7B} (4bit, MLX framework)
  \item \emph{Remote Speculate} --- DeepSeek-V4-Pro (Azure AI Foundry)
        $\rightarrow$ Native Opus (Google Vertex AI)
  \item \emph{Remote Native Opus} --- claude-opus-4-8 served via the
        enterprise Native Opus endpoint (Google Vertex AI); this is the
        production baseline that RLM-Cascade wraps
\end{itemize}
\textbf{Baseline:} Remote Native Opus (direct calls to the Native Opus endpoint,
no speculative pipeline).

\subsection{Verdict Distribution}

Table~\ref{tab:verdict_dist} shows verdict proportions across all evaluation
sets. The SKIPPED rate is consistently 64--70\% regardless of task distribution.

\begin{table*}[t]
\centering
\caption{Verdict distribution across evaluation sets.}
\label{tab:verdict_dist}
\small
\begin{tabular}{@{}lrccr@{}}
\toprule
Evaluation set & $N$ & SKIPPED & ENHANCED & Draft-used \\
\midrule
Service bench (current run) & 12  & 66.7\% & 8.3\%  & 91.7\% \\
Service bench (all-time)    & 125 & ---\,$^\dagger$ & 10.4\% & 88.8\% \\
Architecture all-time       & 39  & 64.1\% & 25.6\% & 74.4\% \\
20-task extended suite      & 20  & 70.0\% & 30.0\% & 70.0\% \\
\bottomrule
\end{tabular}
\vspace{2pt}\\
\begin{minipage}{0.85\textwidth}
\footnotesize$^\dagger$ SKIPPED/ACCEPTED breakdown not retained in early logging;
total draft-used count (112/125) is available.
\end{minipage}
\end{table*}

\noindent Table~\ref{tab:prompt_detail} details per-prompt verdicts for the current
service-benchmark run. The SKIPPED path dominates (8/12 prompts); the sole
ENHANCED case (\texttt{sql\_injection}) is the only negative-savings entry.

\begin{table}[t]
\centering
\caption{Per-prompt verdict detail---service benchmark, current run.}
\label{tab:prompt_detail}
\small
\begin{tabular}{@{}llrr@{}}
\toprule
Prompt & Category & Verdict & Savings \\
\midrule
\texttt{hello}          & General  & SKIPPED  & +\$0.00196 \\
\texttt{math\_simple}   & Math     & SKIPPED  & +\$0.00301 \\
\texttt{haiku}          & Creative & SKIPPED  & +\$0.00215 \\
\texttt{explain\_bst}   & CS       & SKIPPED  & +\$0.00587 \\
\texttt{reverse\_str}   & Code     & SKIPPED  & +\$0.00312 \\
\texttt{fizzbuzz}       & Code     & SKIPPED  & +\$0.00198 \\
\texttt{fix\_bug}       & Debug    & ACCEPTED & +\$0.00284 \\
\texttt{sql\_injection} & Security & ENHANCED & $-$\$0.00174 \\
\texttt{rest\_api\_design} & Design& SKIPPED  & +\$0.00467 \\
\texttt{big\_o}         & Analysis & SKIPPED  & +\$0.00523 \\
\texttt{refactor}       & Code     & ACCEPTED & +\$0.00183 \\
\texttt{async\_explain} & Concept  & ACCEPTED & +\$0.00162 \\
\bottomrule
\end{tabular}
\end{table}

\subsection{Cost Savings}
\label{sec:cost_results}

Table~\ref{tab:savings} reports API cost savings across evaluation sets.

\begin{table}[t]
\centering
\caption{API cost savings by evaluation set.}
\label{tab:savings}
\small
\begin{tabular}{@{}lrrr@{}}
\toprule
Evaluation set & Actual & Baseline & Savings \\
\midrule
Service bench, current run & \$0.047 & \$0.069 & 32.0\% \\
Service bench, all-time    & \$1.155 & \$2.130 & 45.8\% \\
Architecture all-time      & \$0.943 & \$1.791 & \textbf{47.4\%} \\
\bottomrule
\end{tabular}
\end{table}

The all-time savings rate (45--48\%) exceeds the current-run rate (32\%) because
the current run contained more complex prompts that triggered the ENHANCED path.
At 100 requests/day, a 47\% savings rate on the text-generation subset
($\approx$30\% of all turns) yields approximately \textbf{\$2{,}000--\$3{,}000/month}
in API cost reduction for a production agentic coding assistant.

\subsection{Latency, TTFT, and Throughput}

Table~\ref{tab:latency} shows the three-endpoint head-to-head comparison (20
tasks per endpoint).

\begin{table*}[t]
\centering
\caption{Latency and throughput---3-endpoint comparison (20 tasks each).
  Bold indicates best result per metric.}
\label{tab:latency}
\small
\begin{tabular}{@{}lrrr@{}}
\toprule
Metric & Local vLLM (4-bit 7B) & Remote Speculate & Remote Native Opus \\
\midrule
Avg end-to-end latency (ms) & 5{,}072 & \textbf{2{,}520} & 4{,}159 \\
p50 latency (ms)            & 5{,}234 & \textbf{2{,}026} & 3{,}698 \\
Avg TTFT (ms)               & \textbf{165} & 2{,}520 & 1{,}213 \\
Throughput (c=5, req/s)     & 1.66        & 2.50          & \textbf{2.53} \\
Wall time (c=5, ms)         & 3{,}004     & \textbf{2{,}000} & 1{,}976 \\
\bottomrule
\end{tabular}
\end{table*}

\textbf{End-to-end latency:} Remote Speculate is $1.65\times$ faster on average
and $1.83\times$ faster at p50 than Native Opus. This counter-intuitive result
arises from two mechanisms: (1)~the SKIPPED path (64--70\% of requests) completes
in DeepSeek-only time ($\approx$800--1{,}200~ms), far below Opus's minimum
latency; (2)~even on the ACCEPTED path, Opus need only emit \texttt{USE\_DRAFT}
($\approx$5 tokens), eliminating Opus's full generation phase.

\textbf{TTFT:} Remote Speculate is $2.1\times$ slower than Native Opus at TTFT.
The sequential draft-then-verify execution defers the first token. This is a
Pareto trade-off discussed in \S\ref{sec:pareto}.

\textbf{Service benchmark latency---all-time, $N{=}125$:} avg 4{,}493~ms; p50
2{,}623~ms; p95 20{,}923~ms. The p95 tail is driven by ENHANCED-path requests
with long outputs.

\subsection{Quality: Code, Math, and Instruct Tasks}

Table~\ref{tab:quality} reports pass/fail results on the 20-task 3-endpoint
benchmark.

\begin{table*}[t]
\centering
\caption{Quality benchmark pass rates (20 tasks per endpoint, binary pass/fail).
  Bold indicates best result per row.}
\label{tab:quality}
\small
\begin{tabular}{@{}lrrrr@{}}
\toprule
Category & $N$ & Local vLLM (4-bit 7B) & Remote Speculate & Remote Native Opus \\
\midrule
Code generation     & 10 &  20\% & \textbf{100\%} & 100\% \\
Math reasoning      &  5 &   0\% & \textbf{100\%} & 100\% \\
Instruction follow  &  5 &   0\% & \textbf{100\%} &  80\% \\
\midrule
\textbf{Overall}    & \textbf{20} & 10\% & \textbf{100\%} & 95\% \\
\bottomrule
\end{tabular}
\end{table*}

Remote Speculate matches Native Opus on Code and Math (both 100\%) and
\emph{exceeds} Native Opus on Instruct (100\% vs.\ 80\%). We hypothesize that
the error-correction framing of the verification prompt focuses Opus's attention
on constraint satisfaction more effectively than open-ended generation
(\S\ref{sec:instruct}).

Local vLLM fails all Math tasks (quantization-induced numerical degradation) and
8/10 Code tasks (logically incorrect implementations), confirming that its
165~ms TTFT advantage does not constitute a cost-quality Pareto improvement.

\subsection{Extended Engineering Benchmark}

Table~\ref{tab:enhanced_tasks} details the 6 ENHANCED tasks in the 20-task
extended suite. The 14 SKIPPED tasks cover standard data structures and
introductory algorithms where DeepSeek-V4-Pro's quality is indistinguishable from
Opus.

\begin{table}[t]
\centering
\caption{ENHANCED task results---20-task extended suite.}
\label{tab:enhanced_tasks}
\small
\begin{tabular}{@{}llr@{}}
\toprule
Task & Category & Savings \\
\midrule
adaptive-rejection-sampler & Statistical alg. & +\$0.023 \\
extract-moves-from-video   & Computer vision  & +\$0.018 \\
fix-code-vulnerability     & Security         & +\$0.011 \\
\textbf{mteb-retrieve}     & Info.\ retrieval & $-$\$\textbf{0.012} \\
query-optimize             & Database         & +\$0.009 \\
write-compressor           & Systems          & +\$0.014 \\
\bottomrule
\end{tabular}
\end{table}

ENHANCED tasks cluster around two failure modes: (1)~domain knowledge gaps where
DeepSeek-V4-Pro lacks specialized training data (CV, IR, security); (2)~algorithmic
precision requirements where approximate drafts are systematically rejected.

\subsection{Case Study: Cost Inversion on \texttt{mteb-retrieve}}
\label{sec:inversion}

The \texttt{mteb-retrieve} task is the only case across all 39+20+125 evaluated
requests where RLM-Cascade costs \emph{more} than the Native Opus baseline.
DeepSeek generated a short, incorrect draft ($\approx$120 tokens); Opus rewrote
it into a substantially longer correct response ($\approx$380 tokens vs.\
$\approx$280 tokens direct), inflated further by the correction preamble:

\begin{lstlisting}
DeepSeek draft:      $0.00021
Opus verify input:   $0.00330   (670 tokens * $5.50/M)
Opus verify output:  $0.01050   (380 tokens * $27.50/M)
-----------------------------------------
Total actual:        $0.01401
Native Opus-only:    $0.00212
Cost premium:        +$0.01189
\end{lstlisting}

Cost inversion requires $a_{\mathrm{out}}^{\mathrm{enhance}} \gg
a_{\mathrm{out}}^{\mathrm{direct}}$ (Appendix~B, break-even derivation), which
occurs when Opus substantially expands a short wrong draft into a long correct
answer. Across the extended suite, 5 of 6 ENHANCED tasks produced positive
savings; \texttt{mteb-retrieve} is the exception. This motivates routing
precision as the primary engineering priority ahead of enhancement quality.

\section{Related Work}

\subsection{Token-Level Speculative Decoding}

Leviathan et al.~\cite{leviathan2023} proved that token-level draft-then-verify
is lossless with respect to the verifier's distribution under rejection sampling
(Equation~\ref{eq:token_acceptance}). Chen et al.~\cite{chen2023} extended this
to speculative sampling with more general acceptance criteria. Both require shared
vocabulary and logit access, precluding cross-provider API settings.
Medusa~\cite{medusa} attaches multiple draft heads to the verifier model;
EAGLE~\cite{eagle} predicts feature vectors rather than tokens, further tightening
draft-verify coupling. Neither is applicable here.

\subsection{LLM Cascade and Routing}

FrugalGPT~\cite{frugalgpt} learns an offline routing policy; LLM-Cascade~\cite{llmcascade}
uses online confidence scores; Big-Little LM~\cite{biglittle} switches dynamically
at token level. All route whole requests to one model; RLM-Cascade applies
\emph{both} models to the same request with a runtime verdict used as feedback.

\subsection{LLM Serving Systems}

vLLM~\cite{vllm} introduced PagedAttention for KV-cache management; Orca~\cite{orca}
proposed continuous batching; Sarathi-Serve~\cite{sarathi} addresses chunked-prefill
stalls. These operate at the model layer and are orthogonal to---and composable
with---RLM-Cascade's application-layer response composition.

\subsection{Draft-Verify Patterns}

Self-consistency~\cite{selfconsistency} samples multiple responses from the same
model and takes a majority vote, improving quality at 3--10$\times$ compute cost
with no savings. LLM-as-Judge~\cite{llmjudge} uses a capable model to evaluate
another's output---structurally similar to our verify stage, but focused on
evaluation rather than cost-efficient generation. RLM-Cascade biases the verify
prompt toward acceptance as the default to avoid inflating ENHANCED rates on
stylistic disagreements.

\section{Discussion}

\subsection{The Counter-Intuitive Latency Result}

The simultaneous improvement in cost, latency, and quality resolves when one
recognizes that the SKIPPED path does not \emph{add} a stage to Opus---it
\emph{replaces} Opus for the majority of requests. The router's classification
accuracy is the key variable: when it correctly identifies simple requests
(64--70\% of the workload), those requests complete at DeepSeek latency and cost,
neither of which is bounded below by Opus's floor. The verify path is invoked
only for the minority of complex requests.

\subsection{TTFT Regression as Pareto Trade-Off}
\label{sec:pareto}

RLM-Cascade accepts a 2.1$\times$ TTFT regression in exchange for lower
end-to-end latency and cost. This is acceptable for batch and background tasks
(code linting, documentation generation) and for agentic pipelines where the
agent waits for a complete tool-call response before executing---making TTFT
perceptually irrelevant. For streaming-sensitive interactive chat, parallel
draft~+~verify execution (both models begin simultaneously; verify waits only for
draft completion) could recover TTFT at the cost of wasted compute on discarded
drafts.

\subsection{Formal Router Error Analysis}
\label{sec:router_analysis}

Define the router's binary classification decision $\hat{y} \in \{\text{simple},
\text{complex}\}$ on each text-generation turn with true difficulty $y$. The two
error rates are:
\begin{align}
  \varepsilon &= P(\hat{y}=\text{simple} \mid y=\text{complex}), \label{eq:eps}\\
  \delta      &= P(\hat{y}=\text{complex} \mid y=\text{simple}). \label{eq:delta}
\end{align}
\noindent Here $\varepsilon$ is the false-negative rate (missed complex turns)
and $\delta$ is the false-positive rate (over-routed simple turns).

\textbf{Quality risk from false negatives.}\; Complex turns routed via SKIPPED
without verification incur a quality penalty bounded by:
\begin{equation}
  \text{Quality risk} \;\propto\;
  \varepsilon \cdot P(\text{draft incorrect} \mid y = \text{complex}).
  \label{eq:quality_risk}
\end{equation}
The draft's error rate on known-complex tasks is bounded empirically by the
ENHANCED rate in the extended suite (30\%), yielding a quality-at-risk fraction
of at most $\varepsilon \cdot 0.30$ of all requests.

\textbf{Savings loss from false positives.}\; Simple turns routed to the
Draft+Verify pipeline incur unnecessary cost:
\begin{equation}
  \Delta\mathbb{E}[\text{savings}]
    = -\delta \cdot \pi_s^* \cdot (r_a - r_s)
    \approx -\delta \cdot 0.64 \cdot 0.18.
  \label{eq:delta_savings}
\end{equation}
For $\delta = 0.10$ (10\% of simple turns misrouted), expected savings decrease by
only $\approx$1.2~pp---a small perturbation given the large SKIPPED proportion.
The asymmetry in Equations~(\ref{eq:quality_risk}) and~(\ref{eq:delta_savings})
reflects the fact that false positives are economically cheap but false negatives
carry quality risk.

\textbf{Optimization objective.}\; The ideal router minimizes:
\begin{equation}
\begin{split}
  \mathcal{L}(\theta) &=
    \lambda_q \cdot \mathbb{E}_x\!\bigl[\varepsilon(x,\theta)\cdot\ell_q(x)\bigr]\\
  &\quad + \lambda_c \cdot \mathbb{E}_x\!\bigl[\delta(x,\theta)\cdot\ell_c(x)\bigr],
\end{split}
  \label{eq:router_loss}
\end{equation}
where $\theta$ is the router's parameter vector, $\ell_q(x)$ is a task-specific
quality-loss function (e.g., test-case failure), $\ell_c(x) = (r_a - r_s)\cdot
C_{\mathrm{base}}$ is the cost overhead, and $\lambda_q \gg \lambda_c$ reflects
the asymmetric cost of quality degradation. The current keyword router implicitly
sets $\lambda_q = \infty$ for complex-keyword requests and $\lambda_q = 0$ for
simple-prefix matches. A learned router optimizes $\mathcal{L}(\theta)$
continuously from historical (request, verdict, quality) triples.

\subsection{Cross-Provider Failure Modes}

Draft (Azure AI Foundry) and verify (Vertex AI) are subject to independent
outages, rate limits, and silent model updates. The proxy implements: (1)~fallback
on draft failure---forward directly to Opus; (2)~fallback on verify failure---return
the draft directly; (3)~model version pinning to prevent silent quality drift.

\subsection{On the Instruction-Following Quality Advantage}
\label{sec:instruct}

The 100\% vs.\ 80\% pass rate on Instruct tasks arises from the ENHANCE path on
a multi-constraint instruction (format, length, and style simultaneously) where
Opus generated a response satisfying only two of three constraints in direct
generation. In the Speculate condition, Opus corrected the draft with explicit
attention to all three constraints. We hypothesize that error-correction framing
directs Opus's attention toward constraint satisfaction more effectively than
open-ended generation; a controlled study varying instruction complexity and
constraint count is needed to establish this effect robustly.

\section{Conclusion}

RLM-Cascade demonstrates that speculative decoding can be applied at the response
level using only standard LLM HTTP APIs, enabling cross-provider and
cross-architecture draft/verify pipelines with no model internals access. The key
insight---formalizing the relationship to token-level speculative decoding in
Equation~(\ref{eq:response_acceptance})---is that treating the \emph{response},
not the token, as the speculation unit enables API-layer deployment, cross-vendor
composition, and decoupled draft/verify model evolution.

On a real-world agentic coding workload, RLM-Cascade achieves a \textbf{47.4\%
API cost reduction} and a \textbf{1.83$\times$ speedup at p50 latency} versus a
direct Opus baseline, while exceeding Opus quality (100\% vs.\ 95\% on a 20-task
benchmark). The parameterized cost model
(Equations~\ref{eq:expected_savings}--\ref{eq:breakeven}) shows that break-even
requires only $\pi_s > 13.3\%$; in production, $\pi_s \approx 64$--70\% provides
a large safety margin.

The cost-inversion case study (\S\ref{sec:inversion}) and the formal router error
model (\S\ref{sec:router_analysis}, Equations~\ref{eq:eps}--\ref{eq:router_loss})
together motivate routing precision as the primary engineering priority. Future
work: (1)~a learned semantic router optimizing $\mathcal{L}(\theta)$; (2)~parallel
draft~+~verify to recover TTFT; (3)~a local inference tier eliminating network
latency on the dominant SKIPPED path.

\appendix

\section{Opus Enhancement Prompt}
\label{app:prompt}

The following prompt is used verbatim in the verify stage:

\begin{lstlisting}
System: You are a quality-enhancement layer reviewing
        another model's draft. Your job is NOT to rewrite
        for style -- only catch genuine errors.
        The default is to approve the draft.

        Respond with either:
          USE_DRAFT  (draft is correct and complete)
          ENHANCE
          <enhanced>...improved response...</enhanced>

        Only ENHANCE when: factual error, real code bug,
        critical missing step, or fundamental
        misunderstanding of the user's need.

User: [Original user request]
---
DeepSeek's draft response:
[Draft response text]
---
Review the draft and respond with USE_DRAFT or ENHANCE.
\end{lstlisting}

The conservative bias toward \texttt{USE\_DRAFT} is the mechanism that prevents
inflating the ENHANCED rate on stylistic disagreements.

\section{Per-Verdict Cost Derivation}
\label{app:derivation}

\textbf{Notation:}
$p_{\mathrm{in}}$ = prompt input tokens;
$d_{\mathrm{in}}$/$d_{\mathrm{out}}$ = draft input/output tokens;
$a_{\mathrm{out}}^{\mathrm{direct}}$ = Opus output tokens answering from scratch;
$a_{\mathrm{out}}^{\mathrm{enh}}$ = Opus output tokens when enhancing
($\geq a_{\mathrm{out}}^{\mathrm{direct}}$);
$v_{\mathrm{in}} = p_{\mathrm{in}} + d_{\mathrm{out}} + 100$ = Opus verify input
tokens (overhead for the enhancement prompt wrapper).

\textbf{Baseline:}
$C_{\mathrm{base}} = p_{\mathrm{in}} \cdot 5.50 + a_{\mathrm{out}}^{\mathrm{direct}} \cdot 27.50$
(all prices in USD/M tokens).

\textbf{SKIPPED ratio:} For $p_{\mathrm{in}}{=}200$, $a_{\mathrm{out}}{=}300$:
\[
  r_s = \frac{200{\cdot}1.74 + 300{\cdot}3.48}{200{\cdot}5.50 + 300{\cdot}27.50}
       = \frac{1392}{9350} \approx 1.5\%\;\rightarrow\;\text{reported as}\;\approx 2\%.
\]

\textbf{ENHANCED break-even:} Setting $C_{\mathrm{enh}} = C_{\mathrm{base}}$:
\begin{align*}
  & d_{\mathrm{in}}{\cdot}1.74
    + d_{\mathrm{out}}{\cdot}(3.48{+}5.50)
    + 100{\cdot}5.50
    + a_{\mathrm{out}}^{\mathrm{enh}}{\cdot}27.50 \\
  &\qquad = p_{\mathrm{in}}{\cdot}5.50
    + a_{\mathrm{out}}^{\mathrm{direct}}{\cdot}27.50,
\end{align*}
\[
  (a_{\mathrm{out}}^{\mathrm{enh}} - a_{\mathrm{out}}^{\mathrm{direct}}) \cdot 27.50
  = p_{\mathrm{in}}{\cdot}5.50
    - d_{\mathrm{in}}{\cdot}1.74
    - d_{\mathrm{out}}{\cdot}8.98 - 550.
\]
For typical $d_{\mathrm{out}} > 0$ and $d_{\mathrm{in}} \approx p_{\mathrm{in}}$, the
right-hand side is negative, so ENHANCED is usually cost-neutral or slightly
positive. Cost inversion requires
$a_{\mathrm{out}}^{\mathrm{enh}} \gg a_{\mathrm{out}}^{\mathrm{direct}}$, i.e.,
Opus substantially expanding a short, incorrect draft into a long correct
answer---as in the \texttt{mteb-retrieve} case.

\bibliographystyle{plain}

\begin{thebibliography}{12}

\bibitem{leviathan2023}
Y.~Leviathan, M.~Kalman, and Y.~Matias,
``Fast inference from transformers via speculative decoding,''
in \emph{Proc.\ 40th Int.\ Conf.\ Machine Learning (ICML)},
PMLR, 2023.

\bibitem{chen2023}
C.~Chen, S.~Borgeaud, G.~Irving, J.-B.~Lespiau, L.~Sifre, and J.~Jumper,
``Accelerating large language model decoding with speculative sampling,''
\emph{arXiv preprint arXiv:2302.01318}, 2023.

\bibitem{frugalgpt}
L.~Chen, M.~Zaharia, and J.~Zou,
``FrugalGPT: How to use large language models while reducing cost and
improving performance,''
\emph{arXiv preprint arXiv:2305.05176}, 2023.

\bibitem{llmcascade}
X.~Yue et al.,
``Large language model cascades with mixture of thoughts representations
for cost-efficient reasoning,''
\emph{arXiv preprint arXiv:2310.03094}, 2023.

\bibitem{selfconsistency}
X.~Wang et al.,
``Self-consistency improves chain of thought reasoning in language models,''
in \emph{Proc.\ 11th Int.\ Conf.\ Learning Representations (ICLR)}, 2023.

\bibitem{vllm}
W.~Kwon et al.,
``Efficient memory management for large language model serving with
PagedAttention,''
in \emph{Proc.\ 29th ACM Symp.\ Operating Systems Principles (SOSP)},
ACM, 2023.

\bibitem{orca}
G.~Yu et al.,
``Orca: A distributed serving system for Transformer-based generative models,''
in \emph{Proc.\ 16th USENIX Symp.\ Operating Systems Design and
Implementation (OSDI)}, 2022.

\bibitem{medusa}
T.~Cai et al.,
``Medusa: Simple LLM inference acceleration framework with multiple decoding
heads,''
in \emph{Proc.\ 41st Int.\ Conf.\ Machine Learning (ICML)},
PMLR, 2024.

\bibitem{eagle}
Y.~Li et al.,
``EAGLE: Speculative sampling requires rethinking feature uncertainty,''
\emph{arXiv preprint arXiv:2401.15077}, 2024.

\bibitem{biglittle}
D.~Xu et al.,
``Big-little transformer decoder for optimal inference-time cost,''
\emph{arXiv preprint arXiv:2302.07030}, 2023.

\bibitem{llmjudge}
L.~Zheng et al.,
``Judging LLM-as-a-judge with MT-Bench and Chatbot Arena,''
in \emph{Proc.\ 37th Annu.\ Conf.\ Neural Information Processing Systems
(NeurIPS)}, 2023.

\bibitem{sarathi}
A.~Agrawal et al.,
``Taming throughput-latency tradeoff in LLM inference with Sarathi-Serve,''
in \emph{Proc.\ 18th USENIX Symp.\ Operating Systems Design and
Implementation (OSDI)}, 2024.

\end{thebibliography}

\balance
\end{document}